\documentclass[sigconf,screen]{acmart}
\usepackage{amsmath}
\renewcommand\footnotetextcopyrightpermission[1]{} 

\AtBeginDocument{%
  \providecommand\BibTeX{{%
    \normalfont B\kern-0.5em{\scshape i\kern-0.25em b}\kern-0.8em\TeX}}}

\setcopyright{none}
\copyrightyear{none}

\acmConference[]{}{}
\usepackage{xspace}
\newcommand{\titlepaper}[0]{CASET: Complexity Analysis using Simple Execution Traces for CS$*$ submissions}
\newcommand{\titleshort}{CASET\xspace}
\newcommand{\commentit}[1]{}
\begin{document}

\title{\titlepaper}

\author{Aaryen Mehta}
\authornote{Both authors contributed equally to this research.}
\email{ammehta@iitk.ac.in}
\affiliation{%
  \institution{Indian Institute of Technology Kanpur}
  \city{Kanpur, Uttar Pradesh}
  \country{India}
}

\author{Gagan Aryan}
\email{gagan@iitk.ac.in}
\authornotemark[1]
\affiliation{%
  \institution{Indian Institute of Technology Kanpur}
  \city{Kanpur, Uttar Pradesh}
  \country{India}
}


\begin{abstract}
  The most common method to auto-grade a student's submission in a CS1 or a CS2 course is to run it against a pre-defined test suite and compare the results against reference results. However, this technique cannot be used if the correctness of the solution goes beyond simple output, such as the algorithm used to obtain the result. There is no convenient method for the graders to identify the kind of algorithm used in solving a problem. They must read the source code and understand the algorithm implemented and its features, which makes the process tedious.

We propose \titleshort (Complexity Analysis using Simple Execution Traces), a novel tool to analyze
the time complexity of algorithms using dynamic traces and unsupervised machine
learning. \titleshort makes it convenient for tutors to classify the submissions for a
program into time complexity baskets. Thus, tutors can identify the algorithms used
by the submissions without necessarily going through the code written by the
students. \titleshort's analysis can be used to improve grading and provide detailed
feedback for submissions that try to match the results without a proper algorithm,
for example, hard-coding a binary result, pattern-matching the visible or common
inputs. We show the effectiveness of \titleshort by computing the time complexity of many classes of algorithms like sorting, searching and those using dynamic programming paradigm. 
\end{abstract}




\maketitle

\section{Introduction}
In CS1/CS2, the correctness of the solutions of to programming assignments may go beyond producing correct outputs for a set of inputs. In many cases, it is also required to use a specific algorithm to solve a problem, for example a binary search instead of a linear search. In such cases, the graders have to manually go through the submitted code to identify the actual algorithm used in the submission. Apart from being a tedious process, this method is also prone to mistakes by the evaluator. We propose the \titleshort framework to make this evaluation process smooth and effective. At the core of \titleshort is an instrumentation tool Valgrind~\cite{valgrind}, that generates dynamic execution traces for programs on a number of inputs. 

Valgrind is an instrumentation framework that provides a suite of tools that facilitate the building of dynamic analysis tools. Valgrind tools can automatically detect many memory management and threading bugs and profile your programs in detail. It can also be used to build new tools. We analyze these dynamic traces with unsupervised machine learning techniques and efficiently classify them into pre-decided time complexity baskets. The current implementation has been tried and tested on sorting, searching algorithms and a few problems involving dynamic programming~\cite{10.5555/1614191}. We believe that we can extend the same to other algorithms, including those involving data structures like hash, heaps and graphs.

The authors (an instructor and two students who have taken up CS$*$ courses) are painfully aware of the person-hours that go into manually grading the submissions without the presence of an automated tool. This literature describes their efforts to simplify this process and save the graders' time. 

The main contribution of this paper is to propose and experimentally evaluate a deterministic approach to calculate the asymptotic time complexity of an algorithm using dynamic traces. To demonstrate this, we have used the instrumentation framework Valgrind as proof of concept. However, as we detail in the later parts of the Methodology section, it would not be feasible to harness Valgrind to prepare a full-fledged tool deployed in a real-world CS$*$ lab environment because the trace generation setup is highly computationally expensive. 

\section{Related Work}

There have been many articles published that focus on improving the overall teaching and learning experience of CS$*$ courses. The ASSYST system uses a simple form of tracing for counting execution steps to gather performance measurements~\cite{assyst}. This was implemented in an introductory course in which Ada was used as a teaching language. The number of evaluations is calculated later used for complexity analysis. There has been a lot of work done in the area of providing automated feedback for programming assignments. Prutor~\cite{article} is a cloud-based state-of-the-art tutoring platform that helps in providing personalized feedback to individual students. Bob et al.~\cite{10.1145/3287324.3287474} proposed a heat maps-based approach to provide feedback to visually guide student attention to parts of the code that is likely to contain the issue with the submission without giving so much direction effectively the whole answer is given. Sumit et al.~\cite{10.1145/2635868.2635912} proposed a lightweight programming language extension that allows an instructor to define an algorithmic strategy by specifying specific values that should occur during the execution of an implementation. They proposed a dynamic analysis-based approach to test whether a student's program matches the instructor's specification.

ATLAS provides amortized cost analysis of self-adjusting data structures (splay trees, splay heaps, and pairing heaps)\cite{atlas}. Since our main focus is to provide a framework akin to unit tests run in a continuous integration environment to CS$*$ courses, we only focus on elementary data structures and simple algorithms in this paper. 

Traces based on Valgrind~\cite{valgrind} are used by researchers to help students visualize and trace their code~\cite{opt, guo13pytutor}. Our work complements these tools as it targets the graders and help them improve the efficiency and the effectiveness of grading the submissions.

\begin{figure*}[tbp]
    \centering
    \includegraphics[width = 0.9\linewidth]{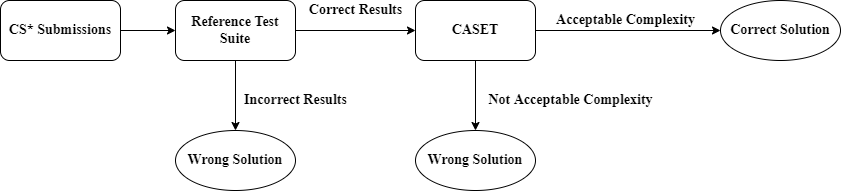}
    \caption{An overview of \titleshort-based pipeline for grading}
    \label{fig:filtering_setup}
\end{figure*}

\section{Methodology}

The submissions for an assignment are run against a pre-defined test suite and compared against reference results. After this, we pass the submission through \titleshort. \titleshort filters out any programs not of the required time complexity. This is similar to unit tests executed within a continuous integration environment. 
Figure~\ref{fig:filtering_setup} shows the grading pipeline setup for \titleshort. 

The primary requirement for \titleshort is the presence of an instrumentation framework for the generation of dynamic traces. We employed Valgrind to demonstrate that time complexity analysis of programs is indeed possible with the presence of dynamic traces. Several experiments were undertaken to test this hypothesis and has been discussed in detail in the Results section.

\commentit{

\begin{figure}[ht]
    \centering
    \includegraphics[width = 1.0\linewidth]{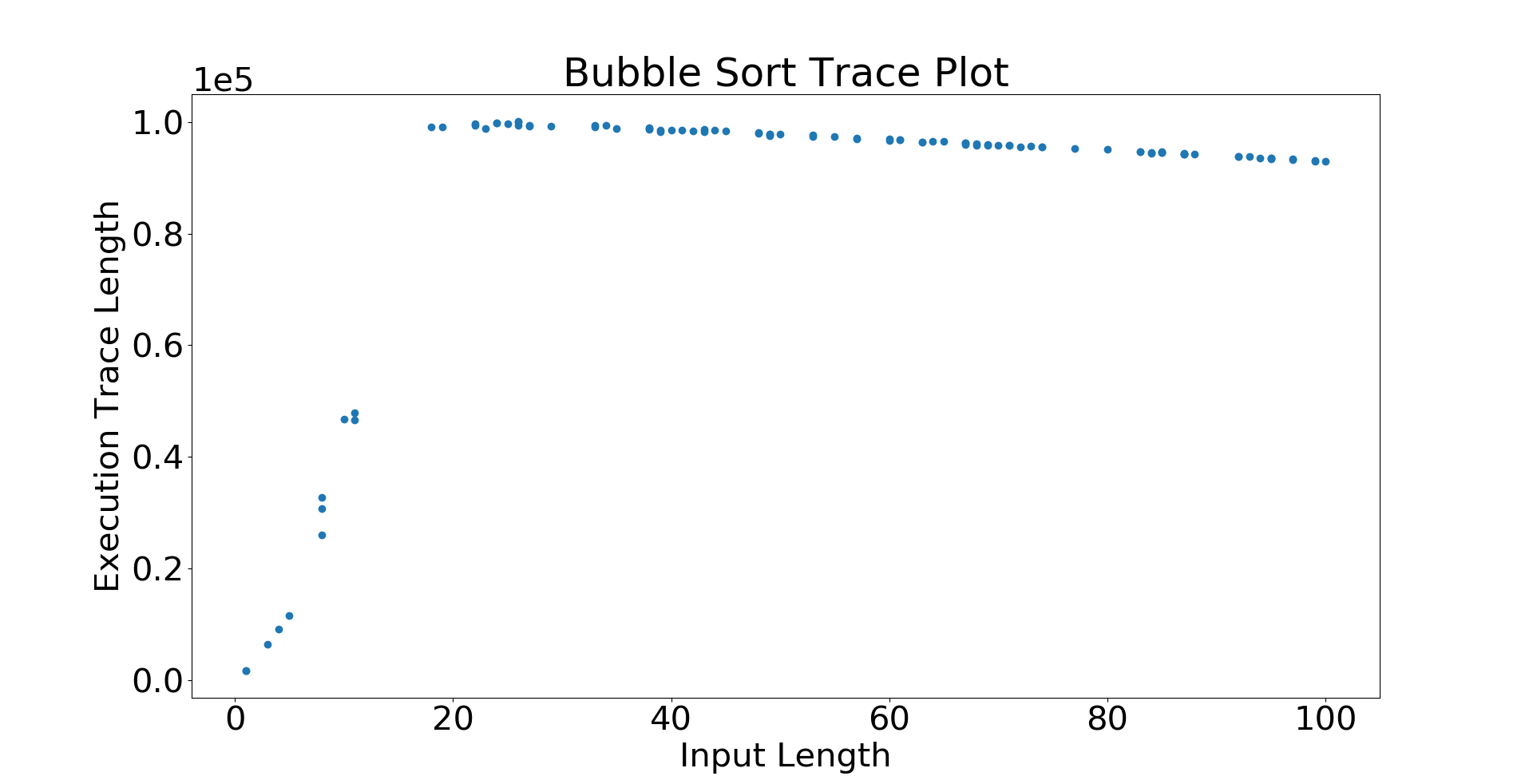}
    \caption{Plot of Valgrind trace length and state change length against input length}
    \label{fig:bubble_iter_1_new}
\end{figure}

Our first experiment was to plot the length of the traces with the input length. We expected that the size of the dynamic traces would increase with increasing input length. However, that wasn't the case. We noticed that the length of the traces decreased with increasing input lengths. (\ref{fig:bubble_iter_1_new})

Upon inspecting the traces, we realized that the generated traces were
truncated. We then began to examine the files in the Valgrind source and noticed a parameter MAX\_STEPS in one of the files used for detecting memory errors. This parameter was added in the backend of Online Python Tutor~\cite{guo13pytutor} because its frontend cannot handle traces of large size(of the order of $10^{5}$). We made relevant changes to bypass this parameter and then generated traces in Valgrind to its maximum capacity.
}

Once the traces are generated, we plot them against the input length of the programs. Scipy's scipy. optimize.curve\_fit is used to estimate the coefficients of the generalized curve equation and plot the curve that fits the best\cite{2020SciPy-NMeth}. Pre-decided curve equations were fit to the traces, and the program was classified to the basket corresponding to which the curve gives the least mean squared error. The curve equations are pre-decided based on the possible time complexities of the algorithm being evaluated. For instance, we will fit the curves $ax + b(\mathcal{O}(n))$ and $\log(ax + b)(\mathcal{O}(\log(n))$ for searching algorithms because these are the only time complexity baskets we expect the algorithm to lie in.

\section{Use Cases}

Let us consider different cases of the performance of programs on a test suite and analyze how \titleshort framework would behave when a program is passed through it. 

\subsection{Case 1 - All the test cases produce the correct result}
When a program produces the same results as the reference results for a test suite, \titleshort will check if the program is implemented in the required time complexity by classifying it to a complexity basket. If the classified basket does not match the time complexity expected from the submission, then the submission would be graded as a wrong solution. 
\subsection{Case 2 - Few of the test cases produce the correct result}
In this case, even though a program produces the correct result for a few test cases, it is possible that the implemented algorithm is not of the required time complexity. So, before assigning a score to the submission, when the program is passed through \titleshort, it computes if the algorithm is of the required time complexity. If it isn't, the submission would be treated as an incorrect solution. \titleshort can also detect hard-coded programs designed to pass a few visible test cases. Hardcoded programs are generally of linear time complexity and wouldn't satisfy the required time complexity (unless the required complexity is linear). This can be detected easily with \titleshort. 

However, it is to be noted that it wouldn't be possible to compute the algorithm's time complexity if it contains runtime, segmentation-fault, or any other memory errors. Because currently, \titleshort uses Valgrind to generate traces, and trace generation from Valgrind is not possible if the program contains memory-related errors. Other instrumentation frameworks that can better handle memory errors should be considered to handle this.

\subsection{Case 3 - None of the test cases produce the correct result}
In this case the program would be graded as incorrect even before passing through \titleshort

\section{Data Analysis}

Valgrind traces can accurately estimate the time complexity of most of the sorting and searching algorithms. We were also able to fit curves on a few dynamic programming algorithms and evaluate their time complexity. The mean squared errors upon plotting different curves with algorithms can be found in Table~\ref{tab:conf}.

\begin{figure*}[p]

\centering
\renewcommand{\arraystretch}{0.98}
\begin{tabular}{cc}
    \includegraphics[width = 0.40\linewidth]{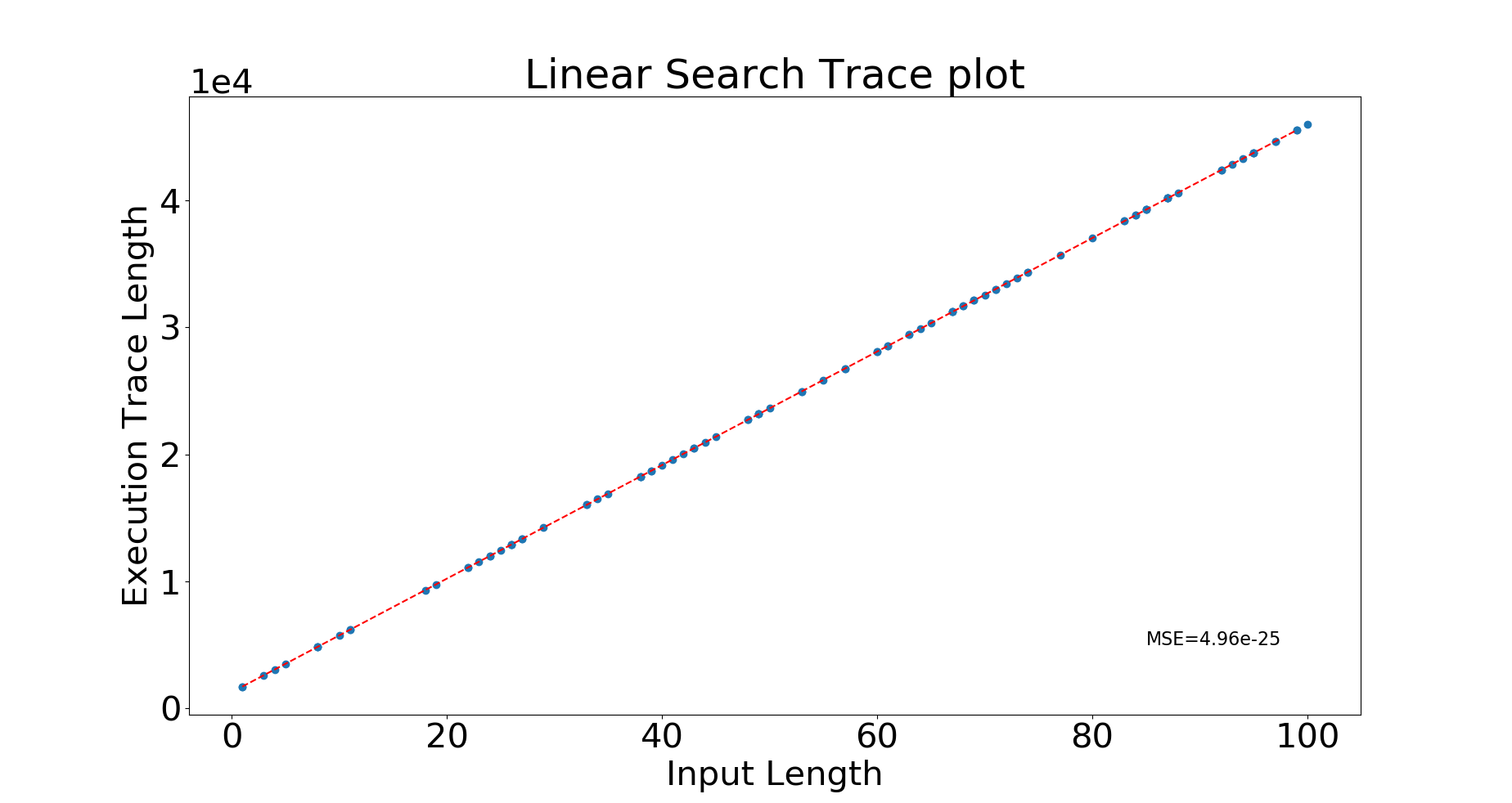} &
    \includegraphics[width = 0.40\linewidth]{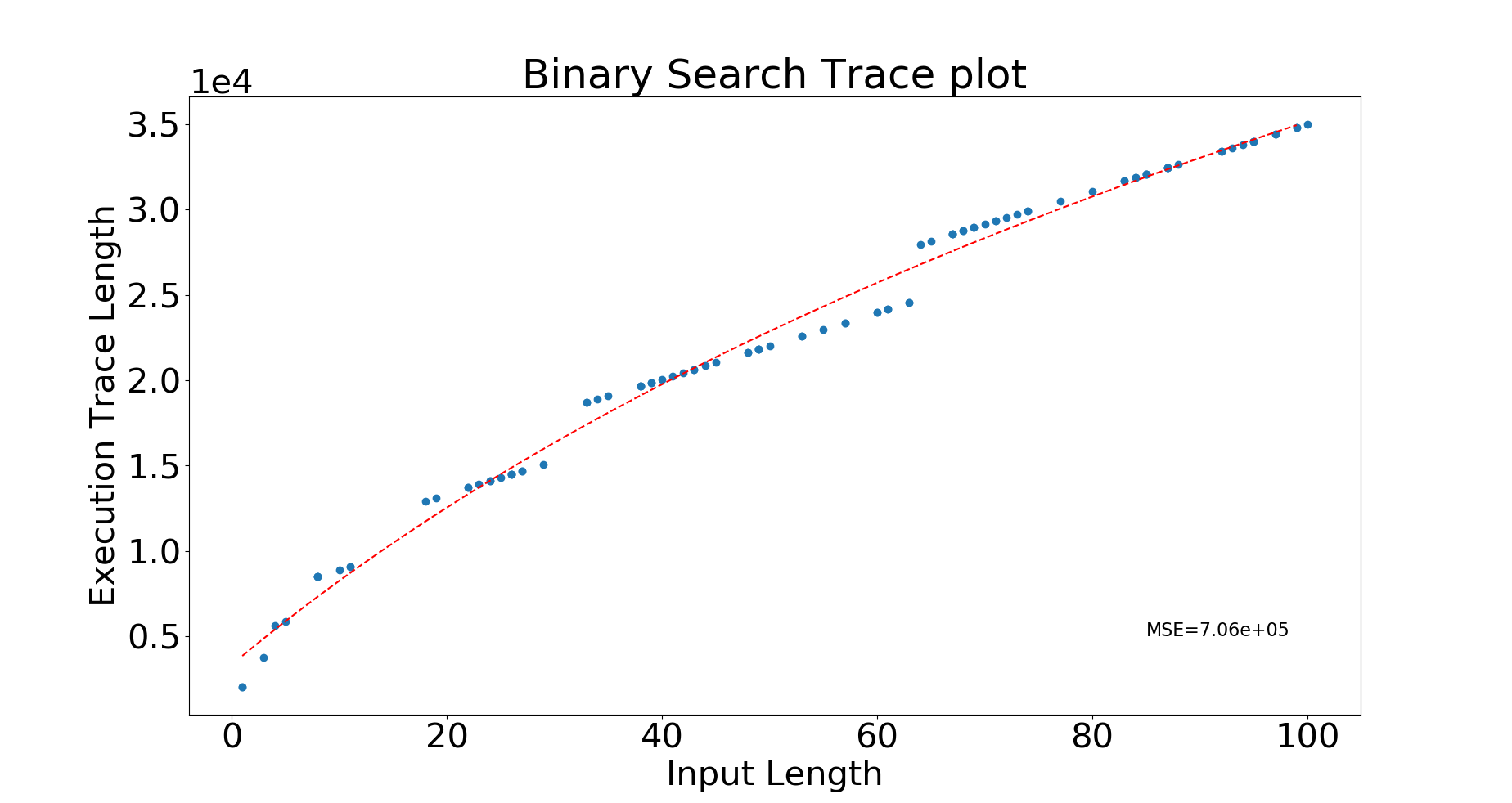} \\
    \textbf{(a) Linear Search; Best Fit Curve: $ax+b$} &
    \textbf{(b) Binary Search; Best Fit Curve: $a\log(x+b)+c$} \\
    \includegraphics[width = 0.40\linewidth]{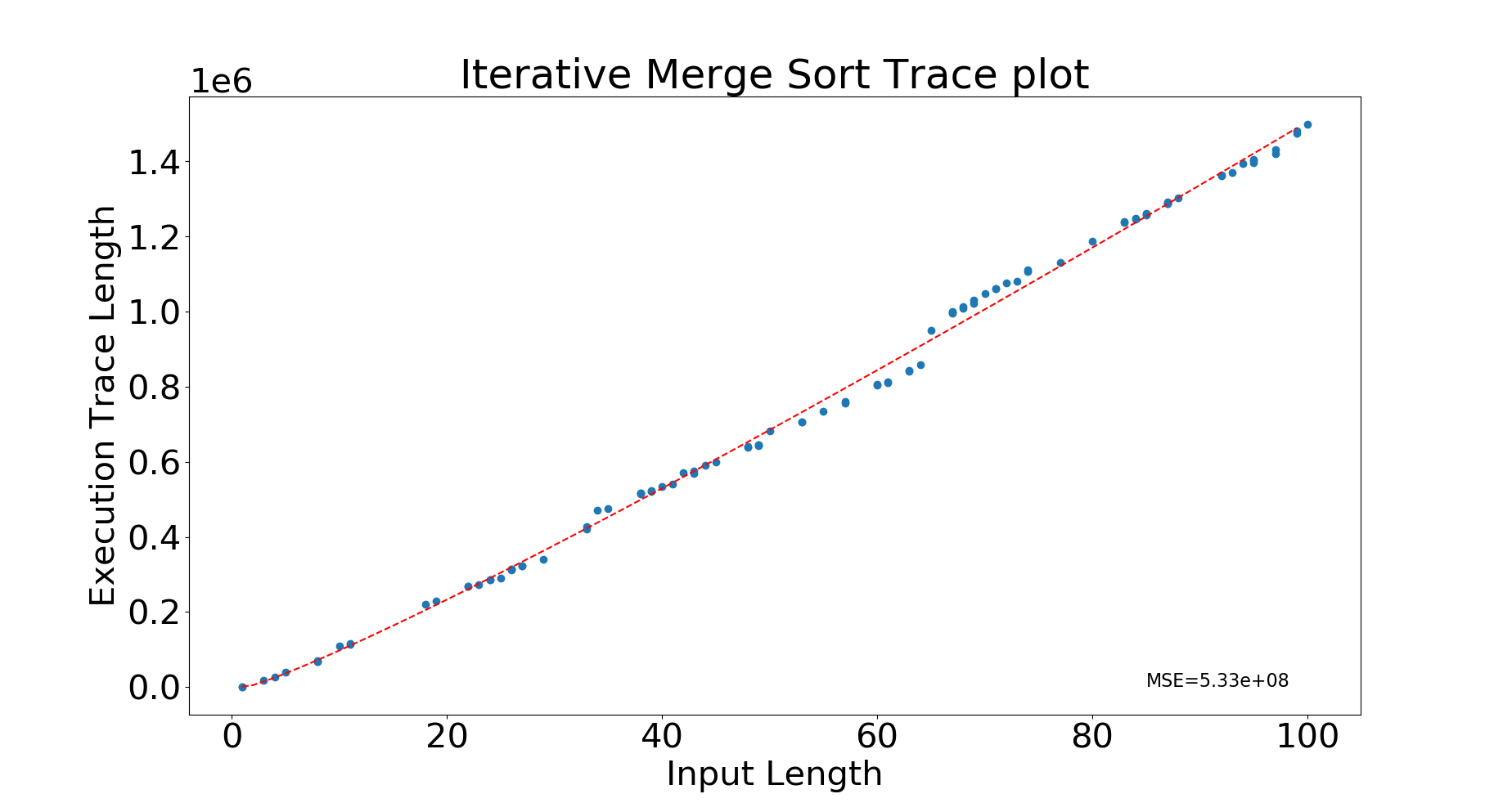} &
    \includegraphics[width = 0.40\linewidth]{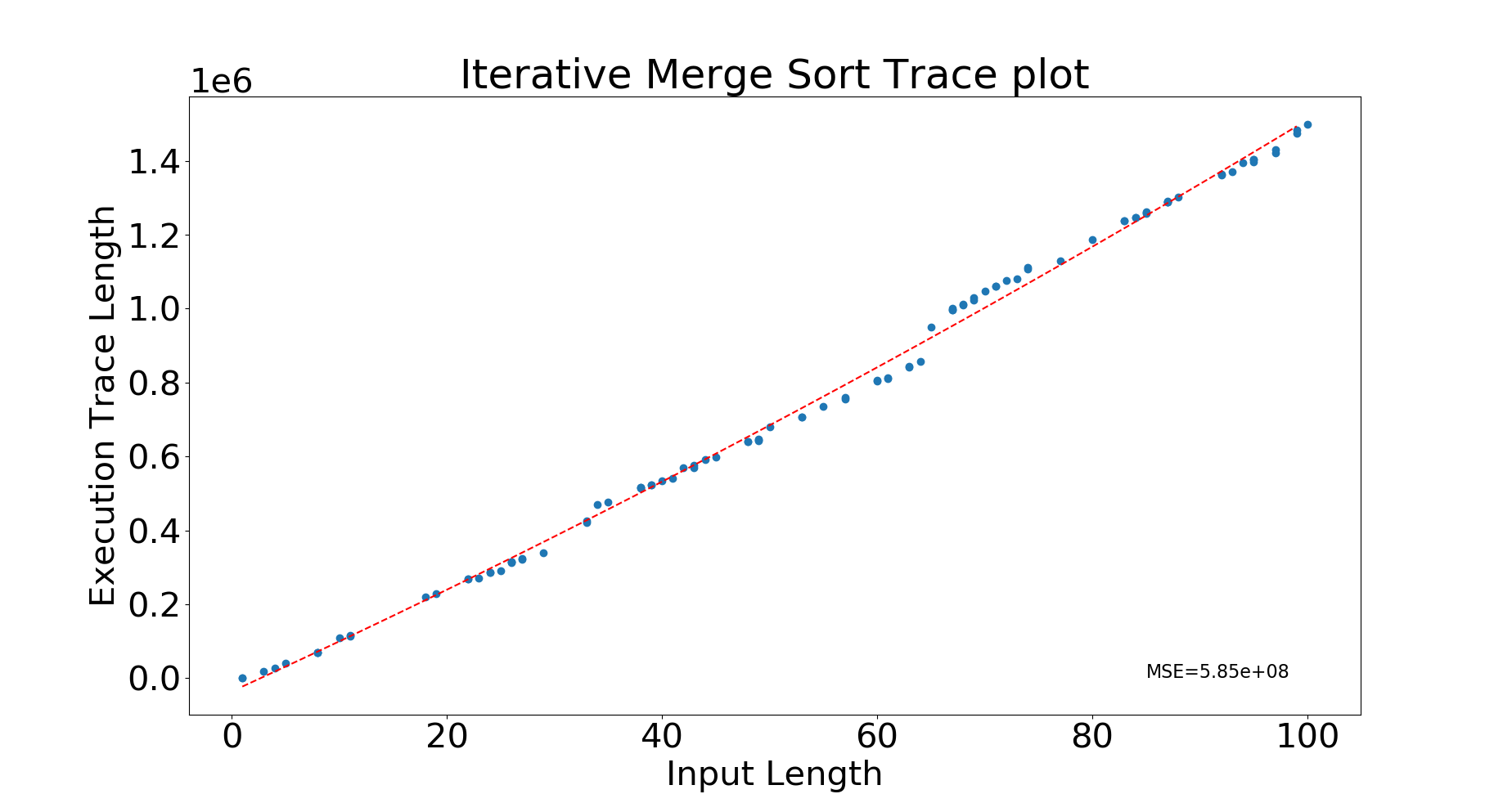} \\
    \textbf{(c) Iterative Merge Sort; Best Fit Curve: $(ax+b)(\log(cx+d))$} & 
    \textbf{(d) Iterative Merge Sort; Fit with Curve: $ax^2+bx+c$} \\
    \includegraphics[width = 0.40\linewidth]{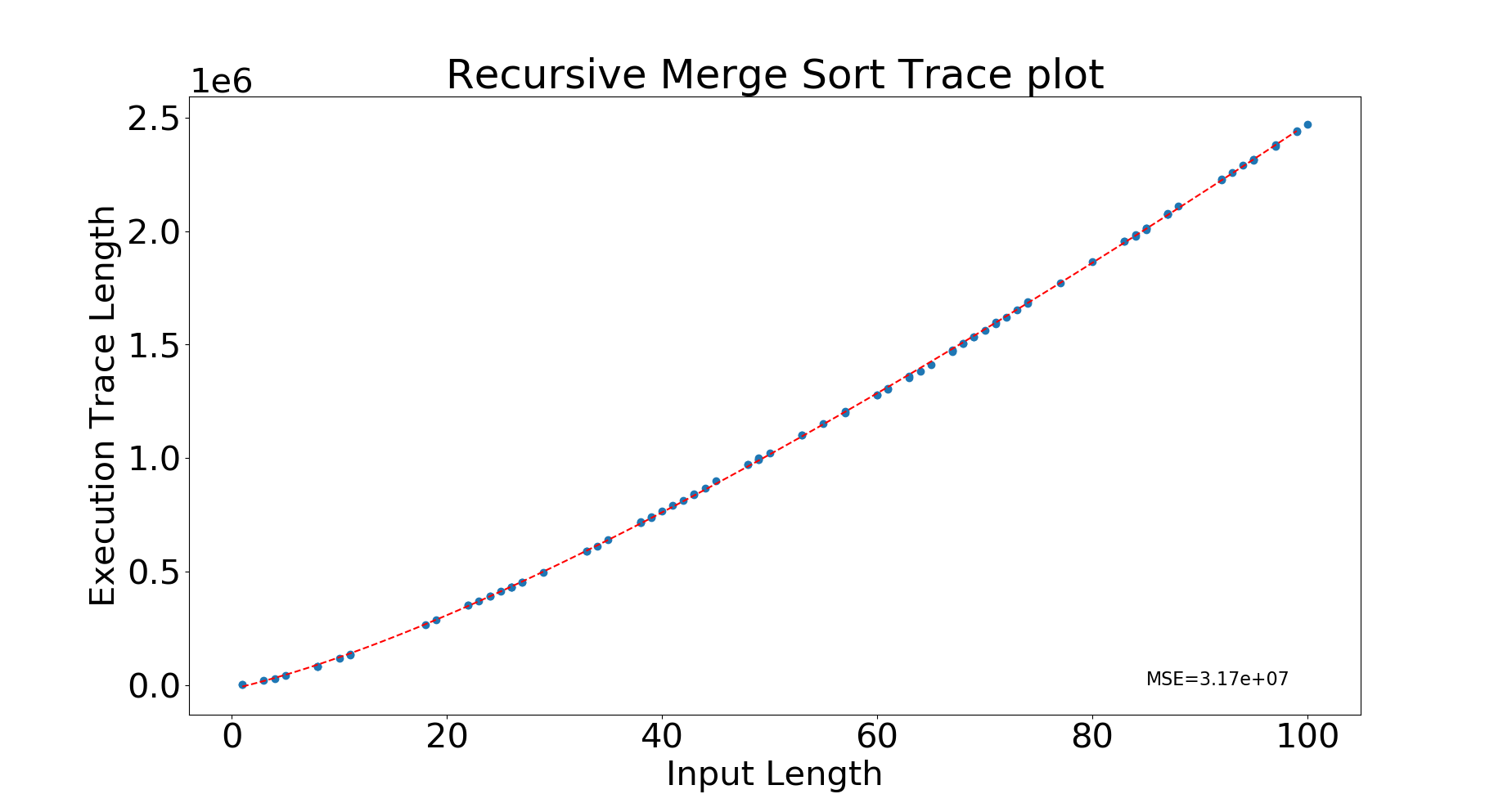} &
    \includegraphics[width = 0.40\linewidth]{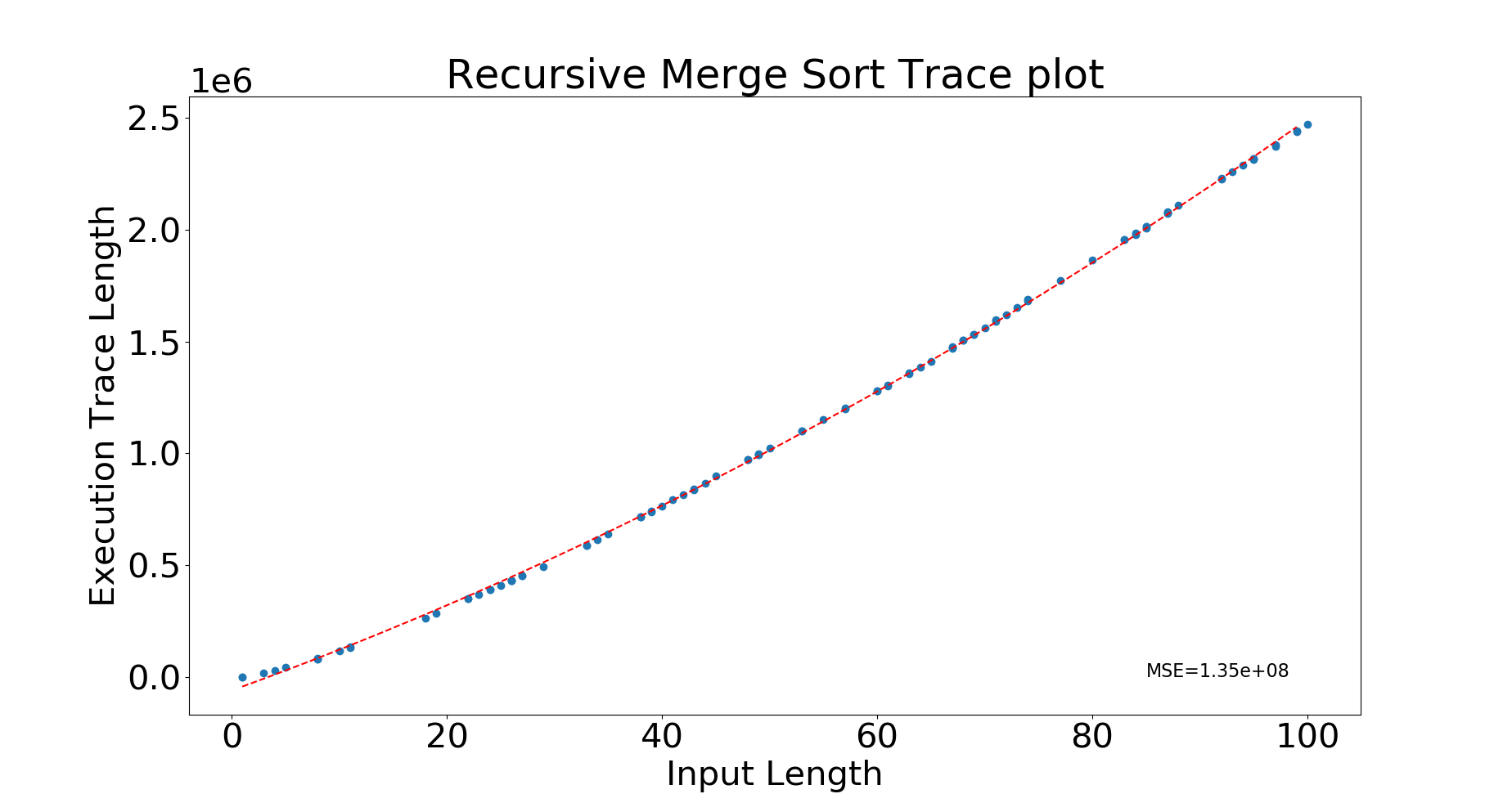} \\
    \textbf{(e) Recursive Merge Sort; Best Fit Curve: $(ax+b)(\log(cx+d))$} &
    \textbf{(f) Recursive Merge Sort; Fit with Curve: $ax^2+bx+c$ } \\
    \includegraphics[width = 0.40\linewidth]{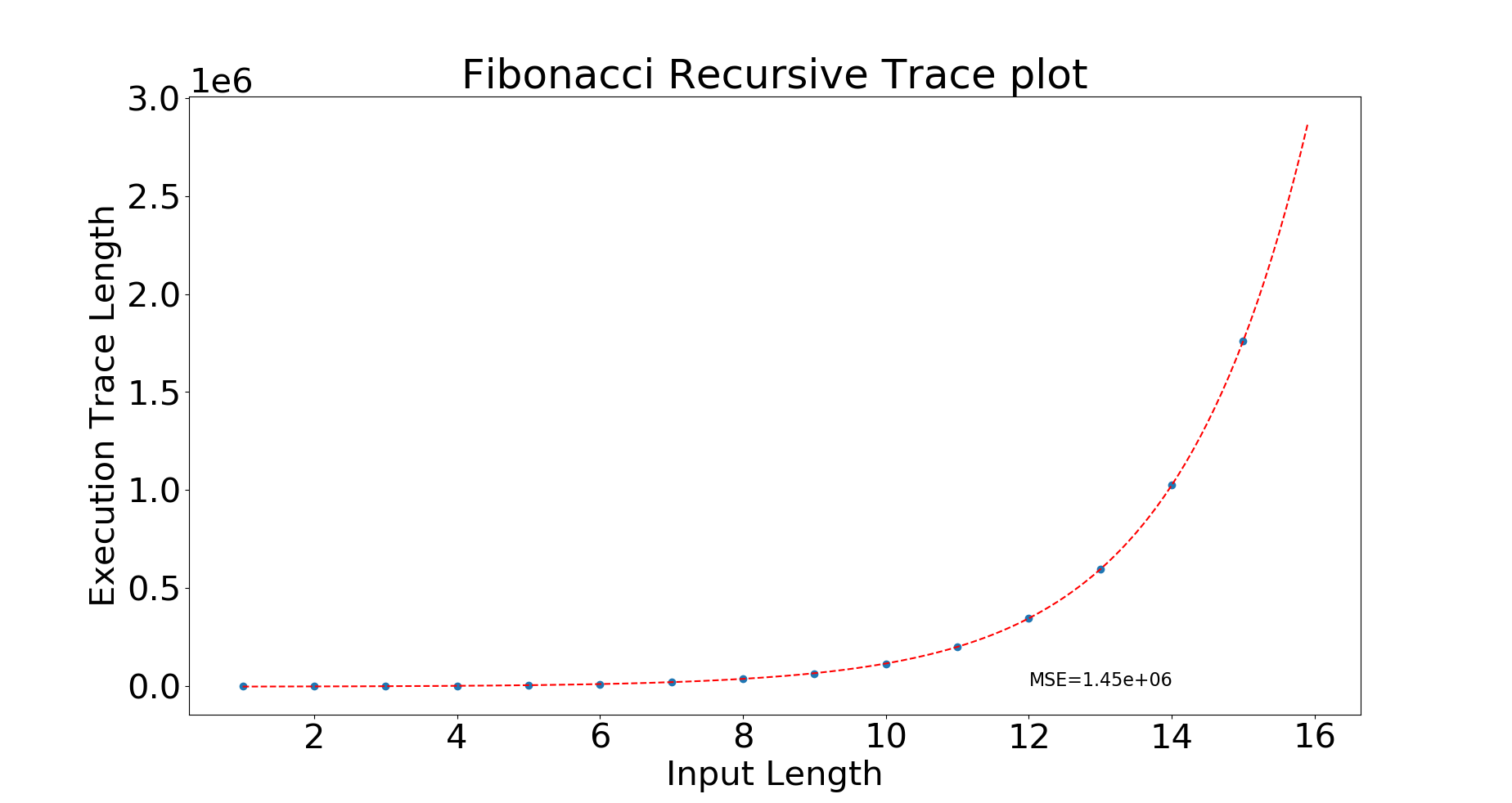} &
    \includegraphics[width = 0.40\linewidth]{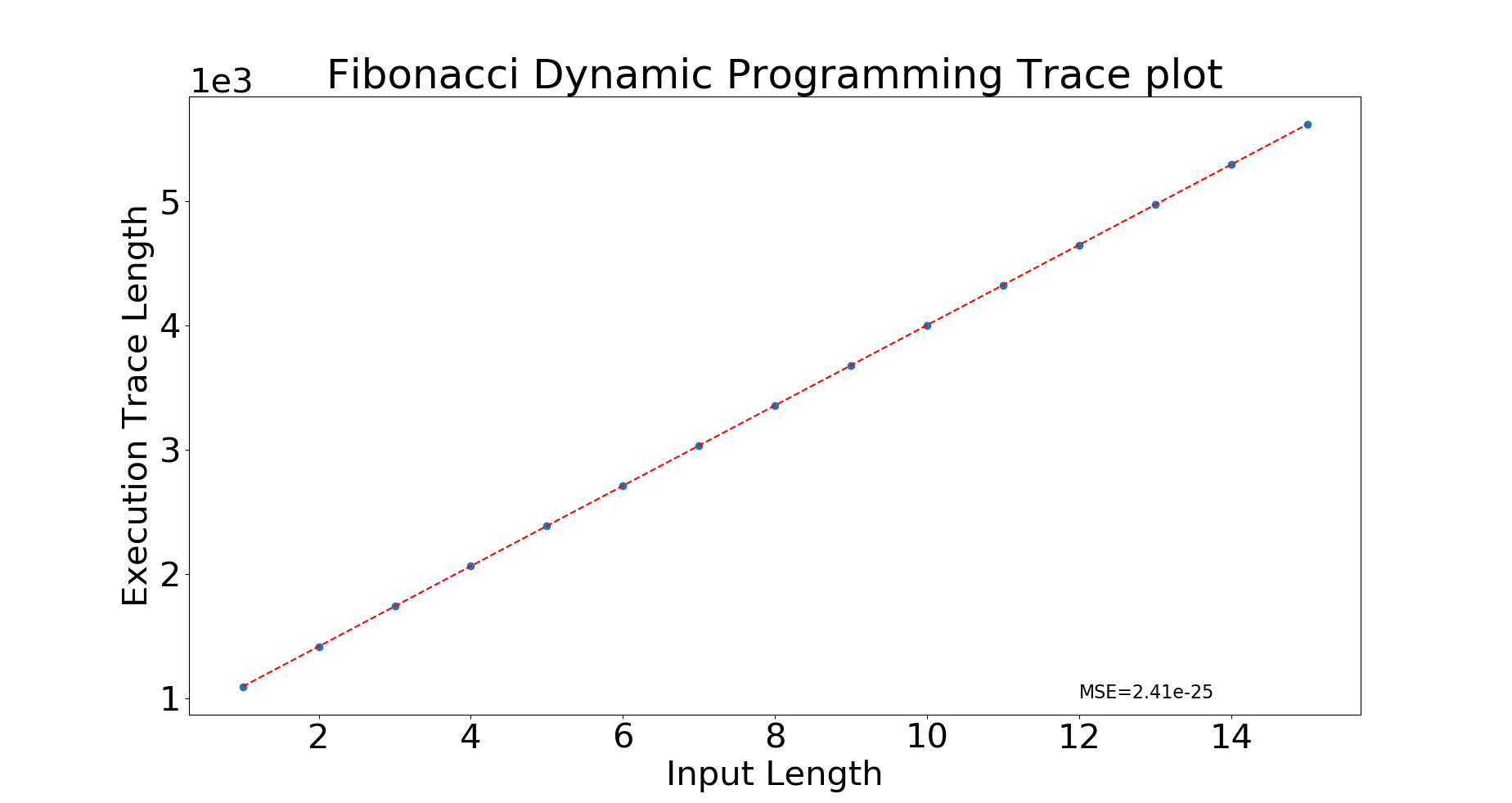} \\
    \textbf{(g) Recursive Fibonacci Algorithm; Best Fit Curve: $e^{ax+b}+c$} &
    \textbf{(h) DP Fibonacci Algorithm; Best Fit Curve: $ax+b$ } \\
    \includegraphics[width = 0.40\linewidth]{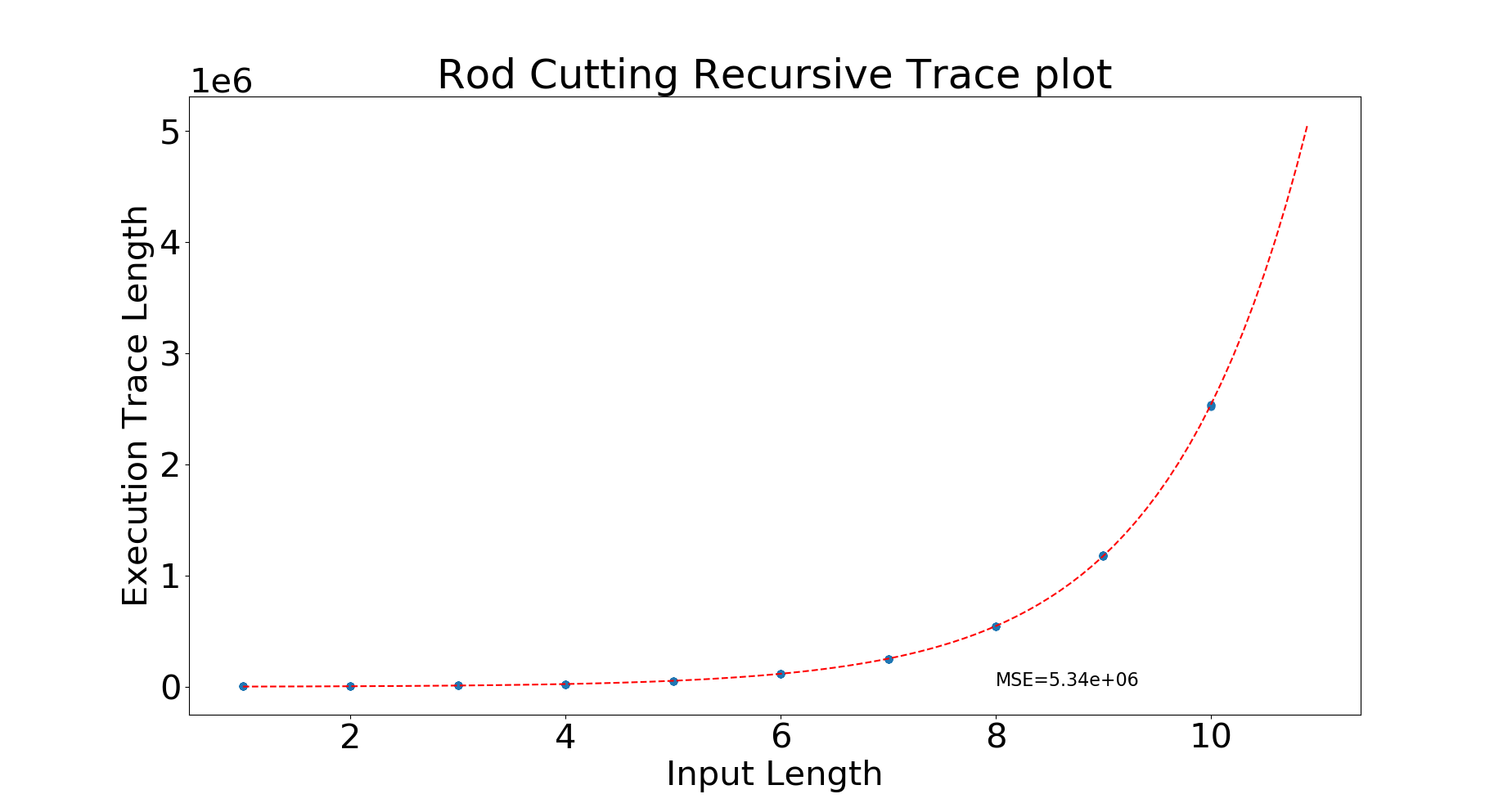} & 
    \includegraphics[width = 0.40\linewidth]{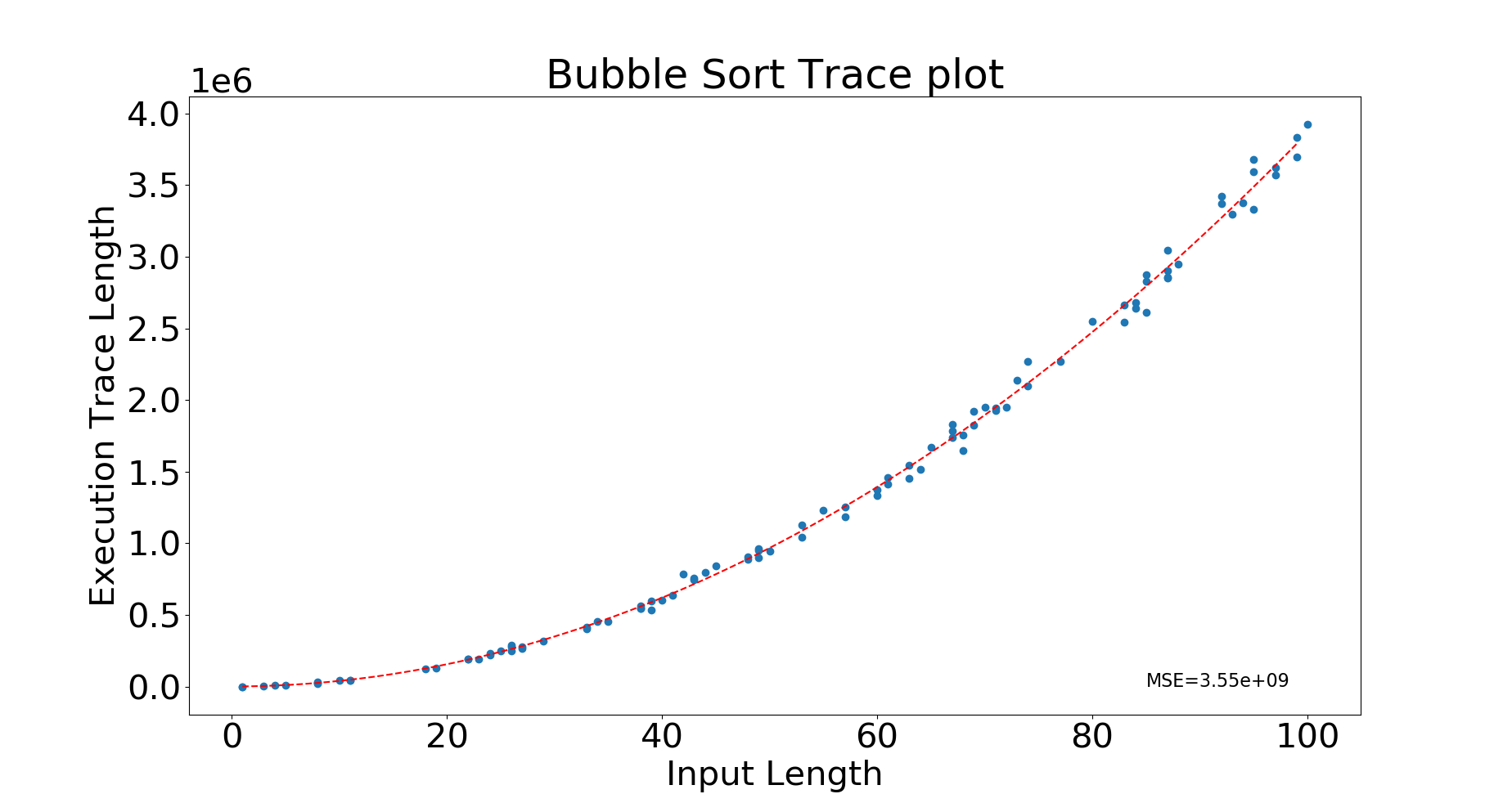} \\
    \textbf{(i) Recursive Rod Cutting Algorithm; Best Fit Curve: $e^{ax+b}+c$} & 
    \textbf{(j) Bubble Sort; Best Fit Curve: $ax^2+bx+c$ } \\
\end{tabular}
\caption{Plots of different algorithms fit with various curves}
\label{tab:graphs}
\end{figure*}

It can be seen in  Table~\ref{tab:conf}  that the algorithms that are of linear time fit the best since their plots are a simple straight line(linear search and dynamic programming fibonacci algorithm), and the best fit can be easily obtained. Even though the other curves seem to fit well in the graphs below, there is a substantial error in the actual curve, and the existing scatter plot, which is not evident due to scale. For instance, even though both $(ax + b)(\log(cx + d))$ and $ax^{2} + bx + c$ seem to fit almost the same in the recursive merge sort plots(plots (e) and (f) in Figure~\ref{tab:graphs}), their mean squared errors differ by a factor of $10$.   

\begin{table*}
\caption{Mean Squared Errors and Optimal Coefficient Values of Curve Fits with Algorithms}
\label{tab:conf}
\begin{tabular}{lllllll}
\toprule
Algorithm & Equation & a & b & c & d & MSE \\
\hline 
Linear Search & $ax + b$ & $4.47 \times 10^2$ &  $1.28 \times 10^3$ & - & - & $4.96 \times 10^{-25}$ \\ 
Linear Search & $a\log(x+b) + c$ & - & - & - & - & NA*\\ %
Binary Search & $ax + b$ & $3.04 \times 10^2$ & $6.56 \times 10^3$ & - & - & $1.80 \times 10^{6}$\\ 
Binary Search & $a\log(x+b) + c$ & $3.37\times10^4$ & $6.36\times10^1$ & $-1.37\times10^5$ & - & $7.06 \times 10^{5}$\\ 
Bubble Sort & $ax^{2} + bx + c$ & $3.85\times10^2$ & $2.19\times10^1$ & $1.89\times10^3$ & - & $3.55 \times 10^{9}$ \\ 
Bubble Sort & $ax + b$ & $4.01\times10^4$ & $-7.33\times10^5$ & - & - & $8.88 \times 10^{10}$ \\ 
Bubble Sort & $(ax + b)(\log(cx + d))$ & - & - & - & - & NA* \\ %
Iterative Merge Sort & $ax + b$ & $1.55\times10^4$ & $-7.68\times10^4$ & - & - & $8.45 \times 10^{8}$ \\ 
Iterative Merge Sort & $(ax + b)(\log(cx + d))$ & $1.70\times10^3$ & $-2.23\times10^3$ & $7.95\times10^1$ & $-7.93\times10^1$ & $5.33 \times 10^{8}$ \\ 
Iterative Merge Sort & $ax^{2} + bx + c$ & $2.13\times10^1$ & $1.33\times10^4$ & $-3.61\times10^4$ & - & $5.85 \times 10^{8}$ \\ 
Recursive Merge Sort & $ax + b$ & $2.58\times10^4$ & $-2.14\times10^5$ & - & - & $3.90 \times 10^{9}$ \\ 
Recursive Merge Sort & $(ax + b)(\log(cx + d))$ & $7.75\times10^3$ & $-1.26\times10^4$ & $2.12\times10^{-1}$ & $4.39\times10^0$ & $3.17 \times 10^{7}$ \\ 
Recursive Merge Sort & $ax^{2} + bx + c$ & $8.11 \times 10^{1}$ & $1.74 \times 10^{4}$ & $-5.95 \times 10^{4}$ & - & $1.35 \times 10^{9}$ 
\\ 
Recursive Fibonacci & $e^{ax + b}+c$ & $5.40 \times 10^{-1}$ & $6.28 \times 10^{0}$ & $-1.85 \times 10^{3}$ & - & $1.45 \times 10^{6}$ 
\\ 
Recursive Fibonacci & $ax + b$ & $8.45 \times 10^{4}$ & $3.96 \times 10^{5}$ & - & - &  $1.01 \times 10^{11}$ 
\\ 
DP Fibonacci & $e^{ax + b}+c$ & - & - & - & - & NA*\\ %
DP Fibonacci & $ax + b$ & $3.23 \times 10^{2}$ & $7.71 \times 10^{0}$ & - & - & $2.41 \times 10^{-25}$ \\ 
DP Rod Cutting Problem & $ax^{2} + bx + c$ & $3.74 \times 10^{2}$ & $7.99 \times 10^{2}$ & $2.11 \times 10^{3}$ & - & $5.68 \times 10^{3}$ 
\\ 
DP Rod Cutting Problem & $e^{ax + b}+c$ & $1.01\times10^1$ & $9.47\times10^0$ & $-1.24 \times 10^{4}$ & - & $1.07 \times 10^{5}$ 
\\ 
Recursive Rod Cutting Problem & $ax^{2} + bx + c$ & $5.07 \times 10^{4}$ & $-3.67 \times 10^{5}$ & $5.09 \times 10^{5}$ & - & $4.17 \times 10^{10}$ 
\\ 
Recursive Rod Cutting Problem & $e^{ax + b}+c$ & $7.67\times10^{-1}$ & $7.072\times10^0$ & $-1.17 \times 10^{3}$ & - & $5.34 \times 10^{6}$ 
\\ 
DP Edit Distance Problem & $e^{ax + b}+c$ & $-6.01 \times 10^1$ & $6.48\times10^0$ & $4.15 \times 10^{5}$ & - & $1.28 \times 10^{11}$ 
\\ 
DP Edit Distance Problem & $ax^{2} + bx + c$ & $1.37 \times 10^{4}$ & $-3.17 \times 10^{4}$ & $6.12 \times 10^{4}$ & - & $5.59 \times 10^{8}$ 
\\ 
Recursive Edit Distance Problem & $e^{ax + b}+c$ & $1.71\times10^0$ & $8.15\times10^0$ & $7.21 \times 10^{2}$ & - & $5.11 \times 10^{5}$ 
\\ 
Recursive Edit Distance Problem & $ax^{2} + bx + c$ & $6.46 \times 10^{5}$ & $2.20 \times 10^{6}$ & $1.67 \times 10^{6}$ & - & $3.99 \times 10^{10}$ \\ 
\bottomrule
\end{tabular}
\footnotesize *scipy.optimize.curve\_fit was not able to produce appropriate parameters for these curves
\end{table*}

\section{Challenges Faced}
Generating dynamic traces of the programs is a computationally expensive task. It was not possible to generate traces for programs even when the size of the array was greater than ten on a machine with an Intel i7 processor and 16GB RAM. So we had to make use of cloud resources. We used Amazon Web Services EC2 instances to generate traces for programs of input lengths greater than 10. Once the generation of full-length dynamic traces of Valgrind was set up, we began to post-process the dynamic traces and fit them into relevant time complexity baskets. The number of inversions in the input array was also taken as a parameter since we only dealt with input length programs < 100.  However, it did not improve the accuracy of the inferences.


Although Valgrind is good enough to demonstrate that the concept behind \titleshort was a valid way to analyze the time complexity of programs, it is not feasible to have Valgrind at the core of \titleshort. The generation of dynamic traces of bubble sort took 2 hours for inputs of length < 100 and 100 test cases. So it is unrealistic to use valgrind for instrumentation in \titleshort in a real-world CS$*$ lab environment, if students expect real time feedback for their submissions.

\section{Conclusion}

In this paper, we have proposed a novel method to determine the asymptotic time complexity of a computer program. The results of this approach on various algorithms show the potential of this approach. However, currently, it is not possible to use them in a CS$*$ lab environment because the trace generation in Valgrind is highly computationally expensive. Other instrumentation frameworks like Dr. Memory~\cite{dmemory} or gperf tools or one built solely for the cause of \titleshort may better fit the requirement of analyzing time complexity. 

However, Valgrind traces demonstrate that the time complexity analysis with this setup is indeed possible. We believe that the same philosophy can be generalized to other algorithms, including those involving data structures like hash, heaps, and graphs. Apart from the time complexity, memory instrumentation can also be used to analyze the space complexity of the program with the presence of an appropriate instrumentation framework. \titleshort (or a framework similar to that) can reduce the pain of the graders by preventing them from going through the program manually. Apart from computing, the time complexity of the submissions, \titleshort can catch the programs that pass a few test cases with hard-coded results. \titleshort can also be a de facto framework to analyze the time complexity of different computer programs. 

\section{Acknowledgement}

This project was conducted under the supervision of Prof. Amey Karkare as part of the Undergraduate Project requirements at IIT Kanpur during Fall 2021 and Spring 2022.


\bibliographystyle{ACM-Reference-Format}
\bibliography{sample-base.bib}


\begin{thebibliography}{11}


\ifx \showCODEN    \undefined \def \showCODEN     #1{\unskip}     \fi
\ifx \showDOI      \undefined \def \showDOI       #1{#1}\fi
\ifx \showISBNx    \undefined \def \showISBNx     #1{\unskip}     \fi
\ifx \showISBNxiii \undefined \def \showISBNxiii  #1{\unskip}     \fi
\ifx \showISSN     \undefined \def \showISSN      #1{\unskip}     \fi
\ifx \showLCCN     \undefined \def \showLCCN      #1{\unskip}     \fi
\ifx \shownote     \undefined \def \shownote      #1{#1}          \fi
\ifx \showarticletitle \undefined \def \showarticletitle #1{#1}   \fi
\ifx \showURL      \undefined \def \showURL       {\relax}        \fi
\providecommand\bibfield[2]{#2}
\providecommand\bibinfo[2]{#2}
\providecommand\natexlab[1]{#1}
\providecommand\showeprint[2][]{arXiv:#2}

\bibitem[Bruening and Zhao(2011)]%
        {dmemory}
\bibfield{author}{\bibinfo{person}{Derek Bruening} {and} \bibinfo{person}{Qin
  Zhao}.} \bibinfo{year}{2011}\natexlab{}.
\newblock \showarticletitle{Practical Memory Checking with Dr. Memory}
  \emph{(\bibinfo{series}{CGO '11})}. \bibinfo{publisher}{IEEE Computer
  Society}, \bibinfo{address}{USA}, \bibinfo{pages}{213–223}.
\newblock
\showISBNx{9781612843568}


\bibitem[Cormen et~al\mbox{.}(2009)]%
        {10.5555/1614191}
\bibfield{author}{\bibinfo{person}{Thomas~H. Cormen},
  \bibinfo{person}{Charles~E. Leiserson}, \bibinfo{person}{Ronald~L. Rivest},
  {and} \bibinfo{person}{Clifford Stein}.} \bibinfo{year}{2009}\natexlab{}.
\newblock \bibinfo{booktitle}{\emph{Introduction to Algorithms, Third Edition}
  (\bibinfo{edition}{3rd} ed.)}.
\newblock \bibinfo{publisher}{The MIT Press}.
\newblock
\showISBNx{0262033844}


\bibitem[Das et~al\mbox{.}(2016)]%
        {article}
\bibfield{author}{\bibinfo{person}{Rajdeep Das}, \bibinfo{person}{Umair Ahmed},
  \bibinfo{person}{Amey Karkare}, {and} \bibinfo{person}{Sumit Gulwani}.}
  \bibinfo{year}{2016}\natexlab{}.
\newblock \showarticletitle{Prutor: A System for Tutoring CS1 and Collecting
  Student Programs for Analysis}.
\newblock  (\bibinfo{date}{08} \bibinfo{year}{2016}).
\newblock


\bibitem[Edmison and Edwards(2019)]%
        {10.1145/3287324.3287474}
\bibfield{author}{\bibinfo{person}{Bob Edmison} {and}
  \bibinfo{person}{Stephen~H. Edwards}.} \bibinfo{year}{2019}\natexlab{}.
\newblock \showarticletitle{Experiences Using Heat Maps to Help Students Find
  Their Bugs: Problems and Solutions}. In \bibinfo{booktitle}{\emph{Proceedings
  of the 50th ACM Technical Symposium on Computer Science Education}}
  (Minneapolis, MN, USA) \emph{(\bibinfo{series}{SIGCSE '19})}.
  \bibinfo{publisher}{Association for Computing Machinery},
  \bibinfo{address}{New York, NY, USA}, \bibinfo{pages}{260–266}.
\newblock
\showISBNx{9781450358903}
\urldef\tempurl%
\url{https://doi.org/10.1145/3287324.3287474}
\showDOI{\tempurl}


\bibitem[Gulwani et~al\mbox{.}(2014)]%
        {10.1145/2635868.2635912}
\bibfield{author}{\bibinfo{person}{Sumit Gulwani}, \bibinfo{person}{Ivan
  Radi\v{c}ek}, {and} \bibinfo{person}{Florian Zuleger}.}
  \bibinfo{year}{2014}\natexlab{}.
\newblock \showarticletitle{Feedback Generation for Performance Problems in
  Introductory Programming Assignments}. In
  \bibinfo{booktitle}{\emph{Proceedings of the 22nd ACM SIGSOFT International
  Symposium on Foundations of Software Engineering}} (Hong Kong, China)
  \emph{(\bibinfo{series}{FSE 2014})}. \bibinfo{publisher}{Association for
  Computing Machinery}, \bibinfo{address}{New York, NY, USA},
  \bibinfo{pages}{41–51}.
\newblock
\showISBNx{9781450330565}
\urldef\tempurl%
\url{https://doi.org/10.1145/2635868.2635912}
\showDOI{\tempurl}


\bibitem[Guo(2013)]%
        {guo13pytutor}
\bibfield{author}{\bibinfo{person}{Philip~J. Guo}.}
  \bibinfo{year}{2013}\natexlab{}.
\newblock \showarticletitle{Online Python Tutor: Embeddable Web-Based Program
  Visualization for Cs Education}. In \bibinfo{booktitle}{\emph{Proceeding of
  the 44th ACM Technical Symposium on Computer Science Education}} (Denver,
  Colorado, USA) \emph{(\bibinfo{series}{SIGCSE '13})}.
  \bibinfo{publisher}{Association for Computing Machinery},
  \bibinfo{address}{New York, NY, USA}, \bibinfo{pages}{579–584}.
\newblock
\showISBNx{9781450318686}
\urldef\tempurl%
\url{https://doi.org/10.1145/2445196.2445368}
\showDOI{\tempurl}


\bibitem[Jackson and Usher(1997)]%
        {assyst}
\bibfield{author}{\bibinfo{person}{David Jackson} {and}
  \bibinfo{person}{Michelle Usher}.} \bibinfo{year}{1997}\natexlab{}.
\newblock \showarticletitle{Grading Student Programs Using ASSYST}
  \emph{(\bibinfo{series}{SIGCSE '97})}. \bibinfo{publisher}{Association for
  Computing Machinery}, \bibinfo{address}{New York, NY, USA},
  \bibinfo{pages}{335–339}.
\newblock
\showISBNx{0897918894}
\urldef\tempurl%
\url{https://doi.org/10.1145/268084.268210}
\showDOI{\tempurl}


\bibitem[Leutgeb et~al\mbox{.}(2021)]%
        {atlas}
\bibfield{author}{\bibinfo{person}{Lorenz Leutgeb}, \bibinfo{person}{Georg
  Moser}, {and} \bibinfo{person}{Florian Zuleger}.}
  \bibinfo{year}{2021}\natexlab{}.
\newblock \showarticletitle{ATLAS: Automated Amortised Complexity Analysis of
  Self-adjusting Data Structures}. In \bibinfo{booktitle}{\emph{Computer Aided
  Verification}}, \bibfield{editor}{\bibinfo{person}{Alexandra Silva} {and}
  \bibinfo{person}{K.~Rustan~M. Leino}} (Eds.). \bibinfo{publisher}{Springer
  International Publishing}, \bibinfo{address}{Cham}, \bibinfo{pages}{99--122}.
\newblock


\bibitem[Nethercote and Seward(2007)]%
        {valgrind}
\bibfield{author}{\bibinfo{person}{Nicholas Nethercote} {and}
  \bibinfo{person}{Julian Seward}.} \bibinfo{year}{2007}\natexlab{}.
\newblock \showarticletitle{Valgrind: A Framework for Heavyweight Dynamic
  Binary Instrumentation} \emph{(\bibinfo{series}{PLDI '07})}.
  \bibinfo{publisher}{Association for Computing Machinery},
  \bibinfo{address}{New York, NY, USA}, \bibinfo{pages}{89–100}.
\newblock
\showISBNx{9781595936332}
\urldef\tempurl%
\url{https://doi.org/10.1145/1250734.1250746}
\showDOI{\tempurl}


\bibitem[Ooghe(2016)]%
        {opt}
\bibfield{author}{\bibinfo{person}{Nicolas Ooghe}.}
  \bibinfo{year}{2016}\natexlab{}.
\newblock \emph{\bibinfo{title}{An online C programming tutor}}.
\newblock \bibinfo{thesistype}{Ph.\,D. Dissertation}. \bibinfo{school}{UCL -
  Ecole polytechnique de Louvain}.
\newblock
\urldef\tempurl%
\url{http://hdl.handle.net/2078.1/thesis:4600}
\showURL{%
\tempurl}


\bibitem[Virtanen et~al\mbox{.}(2020)]%
        {2020SciPy-NMeth}
\bibfield{author}{\bibinfo{person}{Pauli Virtanen}, \bibinfo{person}{Ralf
  Gommers}, \bibinfo{person}{Travis~E. Oliphant}, \bibinfo{person}{Matt
  Haberland}, \bibinfo{person}{Tyler Reddy}, \bibinfo{person}{David
  Cournapeau}, \bibinfo{person}{Evgeni Burovski}, \bibinfo{person}{Pearu
  Peterson}, \bibinfo{person}{Warren Weckesser}, \bibinfo{person}{Jonathan
  Bright}, \bibinfo{person}{St{\'e}fan~J. {van der Walt}},
  \bibinfo{person}{Matthew Brett}, \bibinfo{person}{Joshua Wilson},
  \bibinfo{person}{K.~Jarrod Millman}, \bibinfo{person}{Nikolay Mayorov},
  \bibinfo{person}{Andrew R.~J. Nelson}, \bibinfo{person}{Eric Jones},
  \bibinfo{person}{Robert Kern}, \bibinfo{person}{Eric Larson},
  \bibinfo{person}{C~J Carey}, \bibinfo{person}{{\.I}lhan Polat},
  \bibinfo{person}{Yu Feng}, \bibinfo{person}{Eric~W. Moore},
  \bibinfo{person}{Jake {VanderPlas}}, \bibinfo{person}{Denis Laxalde},
  \bibinfo{person}{Josef Perktold}, \bibinfo{person}{Robert Cimrman},
  \bibinfo{person}{Ian Henriksen}, \bibinfo{person}{E.~A. Quintero},
  \bibinfo{person}{Charles~R. Harris}, \bibinfo{person}{Anne~M. Archibald},
  \bibinfo{person}{Ant{\^o}nio~H. Ribeiro}, \bibinfo{person}{Fabian Pedregosa},
  \bibinfo{person}{Paul {van Mulbregt}}, {and} \bibinfo{person}{{SciPy 1.0
  Contributors}}.} \bibinfo{year}{2020}\natexlab{}.
\newblock \showarticletitle{{{SciPy} 1.0: Fundamental Algorithms for Scientific
  Computing in Python}}.
\newblock \bibinfo{journal}{\emph{Nature Methods}}  \bibinfo{volume}{17}
  (\bibinfo{year}{2020}), \bibinfo{pages}{261--272}.
\newblock
\urldef\tempurl%
\url{https://doi.org/10.1038/s41592-019-0686-2}
\showDOI{\tempurl}


\end{thebibliography}

\end{document}